\documentclass[11pt,letterpaper]{article}

\usepackage{fullpage}
\usepackage{times}
\usepackage{epsfig}
\usepackage{graphicx}
\usepackage{amsmath}
\usepackage{amssymb}
\usepackage{comment}
\usepackage{algorithm}
\usepackage{algorithmicx}
\usepackage{algmatlab} 
\usepackage{algpseudocode} 
\usepackage[pagebackref=true,breaklinks=true,letterpaper=true,colorlinks,bookmarks=false]{hyperref}
\usepackage[usenames,dvipsnames,svgnames,table]{xcolor}

\usepackage{amsthm}
\usepackage{multirow}
\usepackage{subfigure} 

\theoremstyle{plain}

\DeclareMathOperator{\R}{{\mathbb R}}

\newcommand{\ba}{\begin{array}}
\newcommand{\ea}{\end{array}}

\DeclareMathOperator{\Diag}{Diag}
\DeclareMathOperator{\Intf}{Intf}

\begin{document}

\title{TOP-SPIN: \\TOPic discovery via Sparse Principal component INterference\footnote{The work of Martin Tak\'{a}\v{c} was supported by the Centre for
Numerical Algorithms and Intelligent Software (funded by EPSRC grant EP/G036136/1 and the Scottish Funding Council) and by the EPSRC grant EP/I017127/1 (Mathematics for Vast Digital Resources). The work of Peter Richt\'{a}rik was supported by EPSRC grants EP/J020567/1 (Algorithms for Data Simplicity) and EP/I017127/1 (Mathematics for Vast Digital Resources).}}

\author{Martin Tak\'{a}\v{c}$^\sharp$, Selin Damla Ahipa\c{s}ao\u{g}lu$^\ddag$, Ngai-Man Cheung$^\ddag$ and Peter Richt\'{a}rik$^\sharp$\\
$^\sharp$ University of Edinburgh\\
$^\ddag$ Singapore University of Technology and Design
}

\maketitle
 \thispagestyle{empty}

\begin{abstract}
We propose a novel topic discovery algorithm for unlabeled images based on the bag-of-words (BoW) framework. We first extract a dictionary of visual words
and subsequently for each image compute a visual word occurrence histogram. We view these histograms as rows of a large matrix from which we extract sparse principal components (PCs). Each PC identifies a sparse combination of visual words which co-occur frequently in some images but seldom appear in others. Each sparse PC corresponds to a topic, and images whose interference with the PC is high belong to that topic, revealing the common parts possessed by the images. We propose to solve the associated sparse PCA problems using an Alternating Maximization (AM) method, which we modify for purpose of efficiently extracting multiple PCs in a deflation scheme. Our approach  attacks the maximization problem in sparse PCA directly and is scalable to high-dimensional data. Experiments on automatic topic discovery and category prediction demonstrate encouraging performance of our approach.
\end{abstract}


\section{Introduction}\label{sec:intro}

The goal of this paper is to design a method performing the following:

\begin{quote}\emph{Given a database of $n$ images, identify $k$ (not necessarily disjoint) collections  of images, $S_1,\dots,S_k$, with each covering a certain ``topic''.}
\end{quote}

Definition of a topic is not provided, and hence we are looking for an \emph{unsupervised} learning method able to first i) automatically identify the topics from the images, and then to ii) form collections of images belonging to these topics~\cite{Grauman:06, Sivic:08, Bart:08, Kinnunen:10}. 

For instance, consider a database of photos with people, cars and buildings on them (without knowing this). Some photos may contain people and no cars nor buildings, some may have people and cars, some may be photos of buildings unspoiled by cars or people. From the viewpoint of the ``cars'' topic, people and buildings are clutter/background. From the viewpoint of the ``people'' topic, cars and buildings are background and not essential. We would wish to be able to automatically discover these three topics. Note that it may be that people and cars always occur together in an image, while people and buildings also always occur together. In that case the topics which we would wish to discover are ``people and cars'' and ``people and buildings''.

It has recently been demonstrated \cite{Zhang_large-scalesparse} that sparse PCA is able to discover topics in  a database of articles. The approach is applied to a data-matrix $A$ where rows correspond to articles, columns to words and $A_{i,j}$ is equal to the frequency of word $j$ in article $i$. For example,  \cite{Zhang_large-scalesparse} showed that in a NYTimes article dataset,
words associated with the first and second sparse PCs are {\em million, percent, business, company, market, companies} and {\em point, play, team, season, game}, respectively. These words discover two of the most important topics in the articles: business and sports.

One of our contributions is to show that a similar approach can be successfully applied to images. As we shall see, identification of topics in image databases can be performed by extracting \emph{sparse principal components} of a matrix whose rows correspond to \emph{all} images in the database, columns to visual words (obtained by quantization of local descriptors such as SIFT, via clustering), with the $(i,j)$ entry representing the frequency of visual word $j$ in image $i$. Images are subsequently assigned to the identified topics using a simple technique we call \emph{interference}: images whose interference with a PC is high form natural topics.

{\bf Contents:} We start in Section~\ref{sec:lit} by briefly reviewing some of the relevant literature. In Section~\ref{sec:algorithm} we propose and describe TOP-SPIN, an  algorithm for topic discovery. Further, in Section~\ref{sec:SPCA} we provide some background for sparse PCA and present a scalable algorithm for extracting sparse PCs. In Section~\ref{sec:Exp} we provide numerical evidence for the efficacy and efficiency of our approach. Finally, we conclude in Section~\ref{sec:contributions} with a brief summary of our main contributions.

\section{Literature Review} \label{sec:lit}

In the unsupervised visual object categorization problem, we attempt to uncover the category information of an image dataset without relying on any information capturing  image content~\cite{Grauman:06, Sivic:08, Bart:08, Kinnunen:10}.  Unsupervised categorization relieves the burden of human labeling and
removes subjective bias. Grauman and Darrell~\cite{Grauman:06} proposed a graph-based method for unsupervised  object categorization.  In their work, the sets of local feature descriptors extracted from individual database images are graph nodes, while graph edges are weighted by the number of correspondences between images. A spectral clustering algorithm is then applied to the graph's affinity matrix to produce image groupings. Sivic et al.~\cite{Sivic:08} demonstrated unsupervised learning of object hierarchy from datasets of unlabeled images.  In their work, the generative Hierarchical Latent Dirichlet Allocation (hLDA) model, previously used for text analysis~\cite{Blei:2004}, is adapted to the visual domain.  Images are represented by a visual vocabulary of quantized SIFT descriptors.  A ``coarse-to-fine'' description of the images with varying degrees of appearance and spatial localization granularity is proposed to facilitate discovery of visual object class hierarchies. Bart et al.~\cite{Bart:08} also proposed unsupervised learning of visual taxonomies independent of Sivic et al.~\cite{Sivic:08}.  They use a modified nonparametric prior over tree structure of a certain depth~\cite{Blei:2004}.  Their modified model allows to represent several topics at each node in the taxonomy and makes available all topics at every node to facilitate visual taxonomies inference.  Images are represented using space-color histograms. Based on the BoW framework, Kinnunen et al.~\cite{Kinnunen:10} applied the self-organization principle and the Kohonen map to solve unsupervised visual object categorization.

Our work is also related to object recognition.  One important difference is that we do not assume any prior category information: as will be discussed, we discover object categories automatically from the dataset, and the testing images are assigned to these object categories using the same framework.

In object recognition, the use of local descriptors with high degree of invariance has become one of the dominant approaches~\cite{zhang:2007}.  In particular, in the BoW approach, an image is represented by a bag of highly-invariant local feature descriptors (e.g.,~\cite{lowe:1999}).  These local descriptors may be further clustered or quantized into a dictionary of visual words~\cite{Sivic:03}.  A visual word occurrence histogram of an image is used to determine a distance for classification of object categories.  To generate a large dictionary of vocabularies, hierarchical quantization can be used to produce a vocabulary tree with the leaf nodes being the visual words~\cite{Nister:2006}.
A recent work of Naikal et al.~\cite{Nikhil} used Sparse PCA  to select informative visual words to improve object recognition.  Given the prior object category information, they apply Sparse PCA to each object category separately to select informative (more useful) visual words within individual categories.  The union of all the informative visual words selected from individual categories forms the overall refined visual dictionary.
Different from Naikal et al.~\cite{Nikhil}, our work discovers object categorization automatically by applying Sparse PCA in a different way (and with different philosophy).  Also, we propose to perform category prediction by projecting the test-image's occurrence histogram vector directly onto the principal components (PCs) associated with the discovered categories, and this  is different from previously-proposed BoW-based object recognition systems.
We argue that with our approach each PC selects and associates co-occurring visual words that are signatures for a category.  The projection of the test-image's histogram onto a PC quantifies the extent of visual words co-occurrence in the test image, which is useful for predicting the category.

\section{Topic Discovery Algorithm} \label{sec:algorithm}

We propose TOP-SPIN (Algorithm 1), a method for TOPic discovery via Sparse Principal component INterference.

\begin{algorithm}[!h]
\caption{TOP-SPIN}
\label{alg:BOOM}
\begin{algorithmic}
 \State \emph{Input:} $n$ images, $p$=\#visual words, $k$=\#topics, $s$=sparsity
 \State 1. \textbf{Representation:}
 \State \phantom{1.} 1a. Represent each image $i$ by a row vector $h^i\in \R^p$
 \State \phantom{1.} 1b. Compute weight vector $w \in \R^p$, $w\geq0$
 \State 2. \textbf{Extract topics via sparse PCA:}
 \State \phantom{2.} 2a. Form $A = H\Diag(w) \in \R^{n\times p}$, where $H_{i:} = h^i$
 \State \phantom{2.} 2b. Extract $s$-sparse PCs $x^1,\dots,x^k\in \R^p$ from $A$
 \State 3. \textbf{Detect topic images via Interference:}
 \State \phantom{3.} 3a. Choose topic threshold values $\delta_1,\dots, \delta_k>0$
 \State \phantom{3.} 3b. $S_l \leftarrow \{i\;:\; \Intf(i,x^l) > \delta_l\}$, $l=1,\dots,k$
 \State \emph{Output:}  $S_l$ (images associated with topic $l$), $l=1,\dots,k$
\end{algorithmic}
\end{algorithm}

\subsection{Step 1}

In \emph{Step 1a} we utilize the standard Bag of Words (BoW) approach, where for each image  we identify keypoints (e.g., by Maximally Stable Extremal Regions (MSER)), and then find local feature descriptors for them (e.g., by SIFT algorithm; SIFT descriptors are 128-dimensional vectors). We identify a high number of descriptors for each image and then select a random subset and perform clustering, obtaining $p$ cluster centers (``visual words''). Each local descriptor in an image is then substituted by the closest visual word (distances are measured in $L_2$ norm). Therefore, image $i$ can be described by a histogram vector $f^i\in \R^p$ as follows: $f^i_j$ is the number of appearances of visual word $j$ in image $i$. For normalization purposes (e.g., sharpness, size) we instead represent each image $i$ by the \emph{normalized histogram} $h^i = f^i / \sum_j f^i_j$. While in this paper we focus on this particular image representation, our framework also applies to other representations.

Some visual words may be more important than others. For instance, a word appearing in all images with identical frequency is not informative and hence can be excluded from further analysis. In \emph{Step 1b} we associate with each visual word  $j=1,2,\dots,p$ a weight $w_j\geq 0$, forming a vector $w\in \R^p_+$. In the experiments in this paper we work with the Term Frequency Inverse Document
Frequency (TF-IDF) weights \cite{Nister:2006} defined by $w_j = \ln (n/n_j)$, where $n_j = |\{i: h^i_j >0\}|$, i.e.,  the number of images containing visual word $j$. If word $j$ occurs in many images, then $w_j$ is small and vice-versa. However, different weights might be preferable depending on the dataset.

\subsection{Step 2}

In this step we extract $k$ leading sparse principal components (sparse PCs) of the matrix $A = H \Diag(w)$, where the $i$-th row of $H\in \R^{n\times p}$ is $h^i$ and $\Diag(w)$ is the $p\times p$ diagonal matrix with vector $w$ on the diagonal. Various sparse PCA formulations were suggested in the literature. Here we propose the $s$-sparse PC $x^l$ to be obtained as the solution of the following optimization problem:
\begin{equation}\label{eq:SPCA}\text{maximize} \; \|A_l x\|_2^2 \; \text{ subject to } \; \|x\|_2\leq 1,\; \|x\|_0\leq s,\end{equation}
where $\|\cdot\|_2$ is the standard Euclidean norm, $\|x\|_0 = |\{i: x_i \neq 0\}|$ (number of nonzero elements in $x$), and $A_{l+1} = A_l -x^l (x^l)^T$ with $A_1=A$. Further, we propose that \eqref{eq:SPCA} be solved by the simple yet powerful Alternating Maximization (AM) framework presented in \cite{RTD:GPOWER}. The authors of \cite{RTD:GPOWER} provide a source code\footnote{\href{https://code.google.com/p/24am/}{https://code.google.com/p/24am/}} called ``24AM'': the method is  scalable, fast and parallel and can be run on multicore machines, GPUs and clusters. However,  24AM does not implement the solution of a sequence of problems \eqref{eq:SPCA} for $l=1,2,\dots,k$ (deflation techniques for sparse PCA are described in \cite{M08}). A naive approach would be to simply solve \eqref{eq:SPCA} in a loop, forming $A_{l+1}$ from $A_l$ as described above. However, this is not efficient due to the structure and sparsity of the problem. We therefore implement our own multicore version of the method in C++ suitable for the task. Our SPCA solver is three orders of magnitude faster than the Augmented Lagrangian Method (ALM) proposed by \cite{Nikhil} for $p=500$ and its advantage is growing with $p$. More details on SPCA, AM, our modifications of AM and a comparison with ALM are given in Section~\ref{sec:SPCA}.

\subsection{Step 3}

Define the \emph{interference} between PC $x^l$ and image $i$ via
\[ \Intf(i,x^l):= | \sum_{j=1}^p h^i_j w_j x^l_j |.\]
That is, it is the absolute value of the inner/dot product between $x^l$ and $a^i := (h^i_1 w_1,h^i_2 w_2,\dots, h^i_p w_p)^T$ (the $i$-th row of $A$).
It is easy to check that $\Intf(i,x^l)$ is in fact the length of the projection of $a^i$ onto $x^l$: it quantifies the extent that image $i$ contains the visual words associated with PC $x^l$.
In \emph{Step 3b} we define $S_l$ to be the set of images $i$ having large enough interference with $x^l$, where the precise quantitative meaning of ``large enough'' is controlled by the parameter $\delta_l$ chosen in \emph{Step 3a}. This parameter can be chosen as follows. We compute the interferences of all images with $x^l$ and subsequently cluster them into two clusters: ``high'' and ``small''. We then pick $\delta_l$ which separates the two clusters, which leads to topic collections $S_l$ adapted to the data. As we shall see from computational experiments (for instance, see Figure~\ref{fig:bmw:interference}), images having high interference with a PC indeed belong to the same topic/category.

\subsection{An Illustrative Example} 
\begin{figure*}
 \centering
   \includegraphics[width=6.5in]{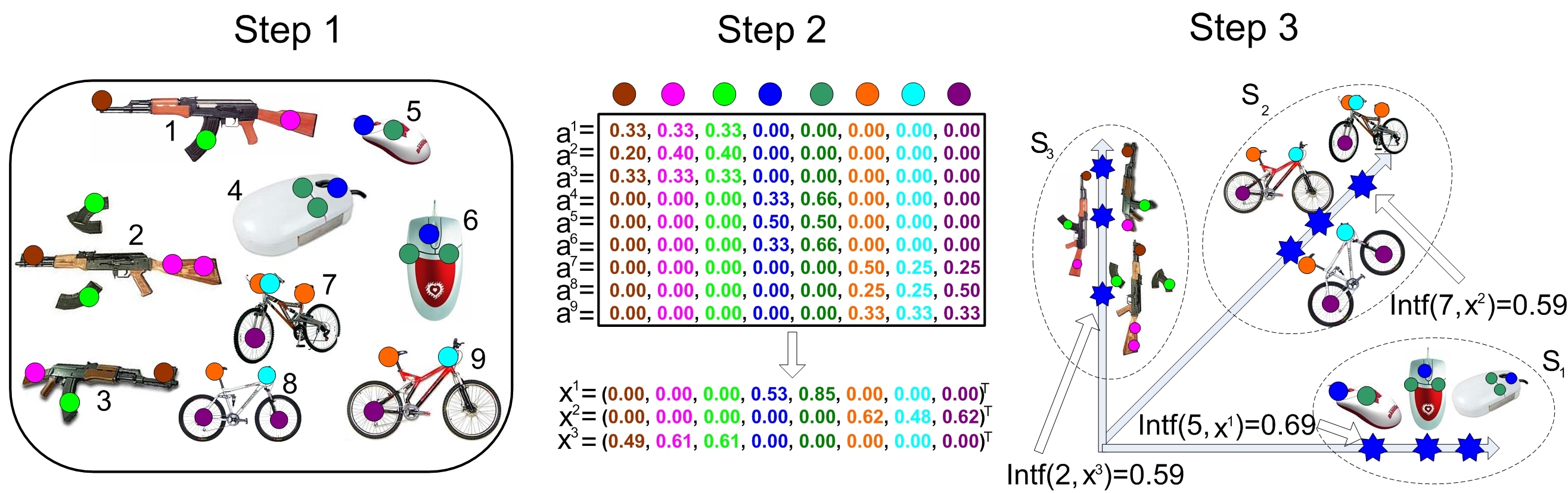}
 \caption{Illustration of the three steps of the TOP-SPIN method. }
 \label{fig:motivation}
\end{figure*}

We illustrate the method on a simplified artificial example (see Figure~\ref{fig:motivation}). We have $n=9$ images which naturally belong to 3 categories/topics: guns, mice and bicycles. In \emph{Step 1} we identify 8 visual words: 3 for guns (green, brown and pink dots), 2 for mice (blue and dark green dots) and  3 for bicycles (light blue, purple and orange dots). In this case the situation is perfect as no two images in different topics contain the same visual word. Here we choose $w$ to be the vector of all ones. As a consequence, $A$ is block diagonal, with rows $a^1, \dots, a^9$ as depicted in \emph{Step~2} in Figure~\ref{fig:motivation}. In \emph{Step~2} of TOP-SPIN, sparse PCs $x^1$, $x^2$ and $x^3$ are computed (we can choose $s=3$). Each sparse PC has zero values outside of two topics and nonzero values in a single topic. In this sense, each sparse PC (perfectly) identifies a topic. In particular, $x^1$ represents the ``mice'' topic, $x^2$ represents the ``bicycles'' topic and $x^3$ represents the ``guns'' topic. Finally, in \emph{Step 3} for each $x^l$ we compute the interferences with each normalized histogram vector $a^i$. The last step in Figure~\ref{fig:motivation} plots each image in a 3D space, with the coordinates of image $i$ being $(\Intf(i,x^1),\Intf(i,x^2),\Intf(i,x^3))$. In this example the interferences of $i$ with $x^l$ will be nonzero if and only if $i$ belongs to the topic represented by PC $x^l$. Hence, each of the sets $S_l$, $l=1,2,3$, will consist of images depicted on a single axis in the 3D space. The three sets $S_1, S_2, S_3$ identified by TOP-SPIN correspond perfectly to the natural topics inherent in the image database.

Real data sets are different from the simplified example depicted in Figure~\ref{fig:motivation} in several ways. First, there will be many images and many visual words. Second, $A$ will not be block diagonal -- images will naturally share visual words with other images since they may share multiple objects. As a consequence, the topics discovered by TOP-SPIN will not be perfect as in the simplified example. Please see Section~\ref{sec:Exp} for numerical experiments with real datasets.

\section{Sparse Principal Component Analysis} \label{sec:SPCA}


Principal Component Analysis (PCA) is an important tool for dimension reduction and data analysis. Let $A \in \R^{n\times p}$ denote a data matrix where the rows correspond to measurements of $p$ variables. PCA finds linear combinations of the columns of $A$, called principal components (PCs), pointing in mutually orthogonal directions, together explaining as much variance in the data as possible. If the rows of $A$ are centered, the problem of extracting the first PC can be written as $\max\{ \| A x \| : \| x\|_2 \leq 1\}$, where $\| \cdot \|$ is any norm for measuring variance\footnote{A simple scaling argument shows that the solution must satisfy $\|x\|_2= 1$.}. Although classical PCA employs the $L_2$ norm, $L_1$ norm can also be used -- this is especially useful when the data is contaminated (e.g., by outliers). Further PCs can be obtained by deflation as explained in the previous section.

PCA usually produces PCs that are combinations of \emph{all} variables. In many applications however, including topic discovery, it is desirable to induce sparsity into the PCs. The problem of finding PCs with few nonzero components is known as sparse PCA or SPCA (see \cite{ABG08}, \cite{AEJL07}, \cite{JNRS10}, and \cite{ZHT04}). Sparsity is usually incorporated either directly enforcing a constraint on the number of nonzero components in a PC, such as in \eqref{eq:SPCA}, or by adding a penalty term to the objective function.

\subsection{SPCA via Alternating Maximization}

We use the open-source 24AM framework  \cite{RTD:GPOWER} for solving the SPCA problem. 24AM is a unifying Alternating Maximization framework for large scale PCA and SPCA problems capable of solving various formulations of SPCA. It also includes parallel implementations of the method for various architectures.
In particular, we find that the  cardinality-constrained formulation \eqref{eq:SPCA} works best, and hence we present 24AM for that case only: Algorithm~\ref{alg:24AM}. The method's name comes from the fact that, for a certain function $F(x,y)$ and convex sets $X$ and $Y$, the two steps of 24AM are of the following alternating maximization form \cite{RTD:GPOWER}: $y = \arg \max_y \{F(x,y): y\in Y\}$ and $x = \arg \max_x \{F(x,y) : x \in X\}$.

\begin{algorithm}[h!]
\caption{24AM: Alternating Maximization}
\label{alg:24AM}
\begin{algorithmic}
 \State Select initial point $x^{(0)}\in \R^p$ and $t \gets 0$
 \State \textbf{Repeat}
 \State \qquad $y^{(t)} = A x^{(t)}/ \|A x^{(t)}\|_2$
 \State \qquad $x^{(t+1)} \gets T_s(A^T y^{(t)})/\|T_s(A^T y^{(t)})\|_2$
 \State \qquad $t \gets t+1$
 \State \textbf{Until} a stopping criterion is satisfied
\end{algorithmic}
\end{algorithm}

By $T_s(a)$ we denote the vector obtained from $a$ by keeping the $s$ largest elements $a_j$ in absolute value and setting the rest to zero.

\subsection{24AM vs ALM}

ALM is an Augmented Lagrangian Method  proposed in \cite{Nikhil} for object recognition and applied to an SDP \emph{relaxation} of the the Sparse PCA formulation in \cite{AEJL07}. On the other hand, 24AM works with the most natural formulation of sparse PCA \emph{directly}. ALM does not control the sparsity level of the solution directly, but via a penalty parameter the value of which is a very poor predictor of sparsity. If a particular target sparsity is sought, one needs to run ALM repeatedly with different values of the penalty parameter, effectively fine-tuning for it. On the other hand, 24AM does not suffer from this issue as sparsity is controlled directly by $s$.

\begin{figure}[H]
\centering
 \includegraphics[width=2.5in]{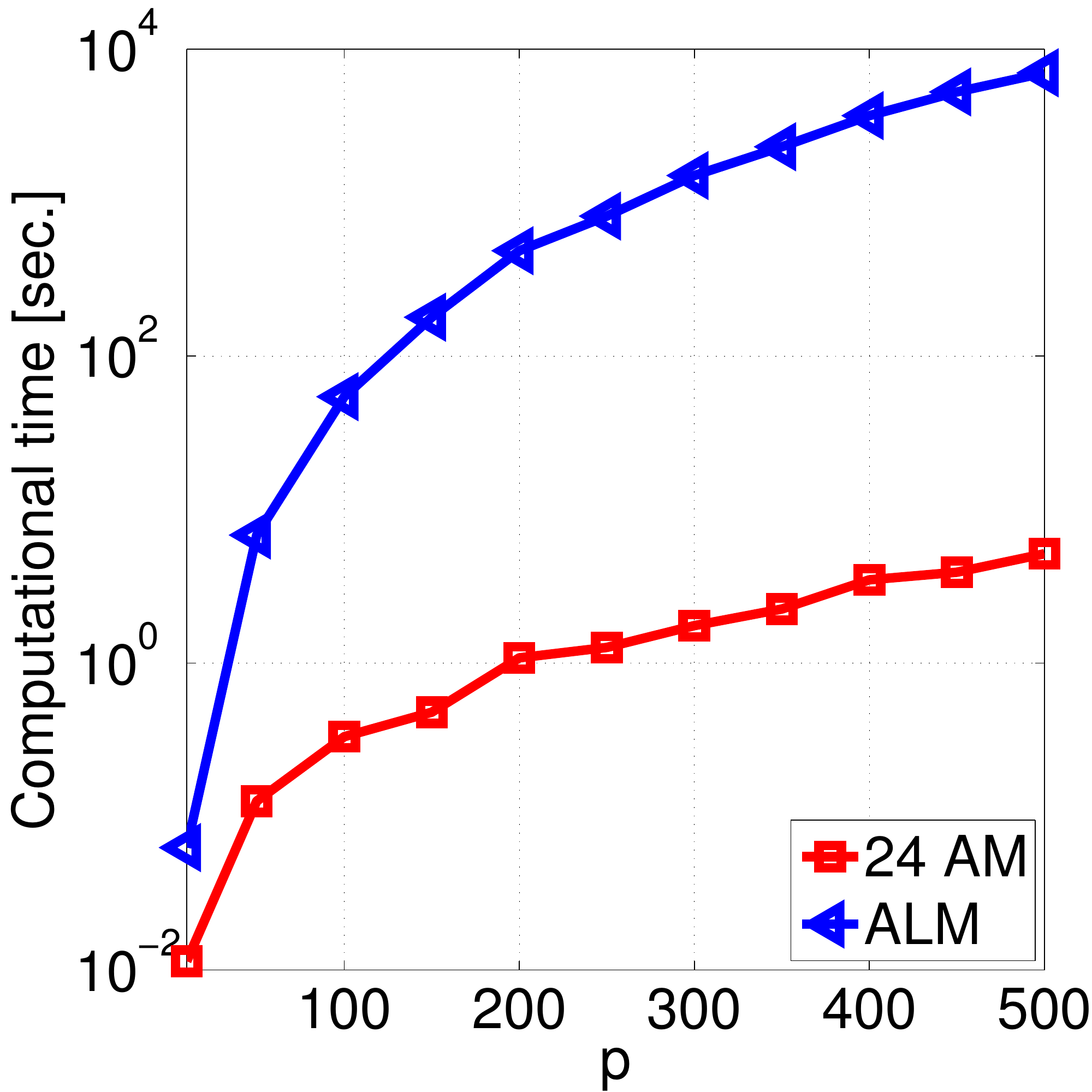}
 \includegraphics[width=2.5in]{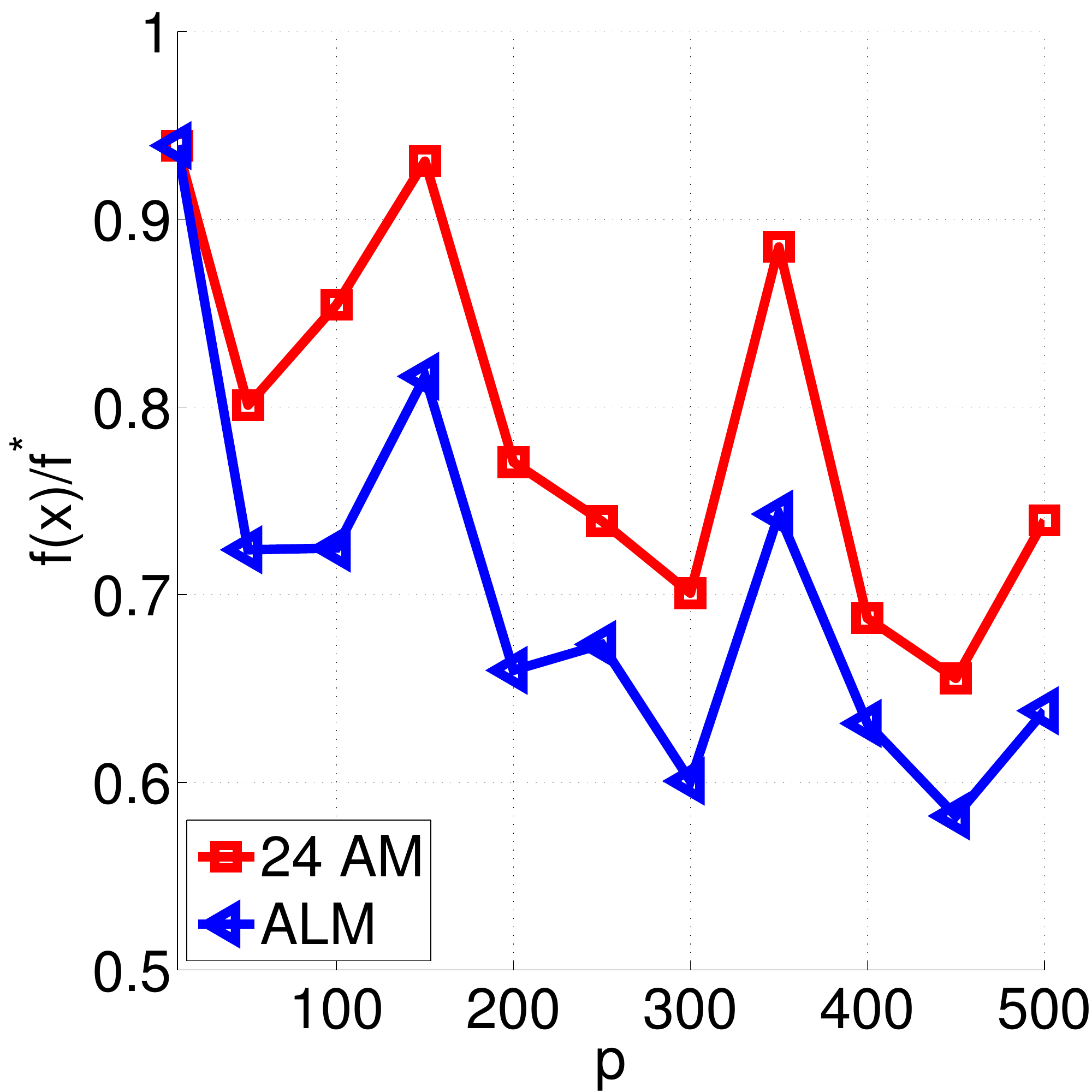}
\caption{24AM vs ALM.}
\label{fig:alm}
\end{figure}

In Figure~\ref{fig:alm} we compare the performance of 24AM and ALM on artificial random matrices $A \in\R^{n\times p}$ with $n=\tfrac p2$ and $p\in\{10, 50, \dots, 500\}$. For each problem we fixed a penalty parameter and obtained a single leading sparse PC using the ALM method. We then measured the resulting sparsity $s$ of the solution. Subsequently, we run 24AM with target sparsity level set to $s$. Here are our findings. First, 24AM terminates \emph{three orders of magnitude faster} than ALM for $p=500$; with the gap getting \emph{larger} with $p$ (left plot). Hence, 24AM is well suited for problems where it is beneficial to work with a large number of visual words. Second, 24AM solutions for all problem instances are of \emph{better quality} than those obtained by ALM (right plot) in the sense that they explain more of the optimal variance. That is, the ratio $f(x)/f^*$ is larger, where $f^* = \|Ax^*\|_2^2$ and $x^*$ is the optimal non-sparse PC, and $f(x) = \|Ax\|_2^2$ and $x$ is the $s$-sparse PC found by the methods.

\section{Numerical Experiments}\label{sec:Exp}

In this section we highlight, on a sequence of carefully chosen experiments, the efficacy and efficiency of TOP-SPIN. We work with the BMW (Berkeley Multiview Wireless) dataset \cite{Fusion, Nikhil} consisting of 20 image categories (Berkeley campus buildings), with $16\times 5 = 80$ images in each. In each category the same building is captured repeatedly from different distances and angles 16 times, each time simultaneously by 5 cameras attached to a fixed frame in close proximity to one another. Hence, there is a total of $1600 = 20\times 16 \times 5$ images.

In all experiments we use MSER keypoints and SIFT descriptors, our codes were implemented in C++. We used OpenCV library v2.4.4.0 to find the keypoints, extract local descriptors and for hierarchical clustering (using FLANN) to obtain a dictionary of visual words.

\subsection{Topic Discovery} \label{sec:pvcForCat}

In this section we empirically show that sparse PCs can identify topics. We took all images from camera \#1 (320 images), used $p=5,000$ visual words
and extracted PCs with $s=20$.

\begin{figure}
 \centering
 \begin{tabular}{cccc}
   & $x^2$ & $x^6$  
   \\
 &
 \includegraphics[width=2in]{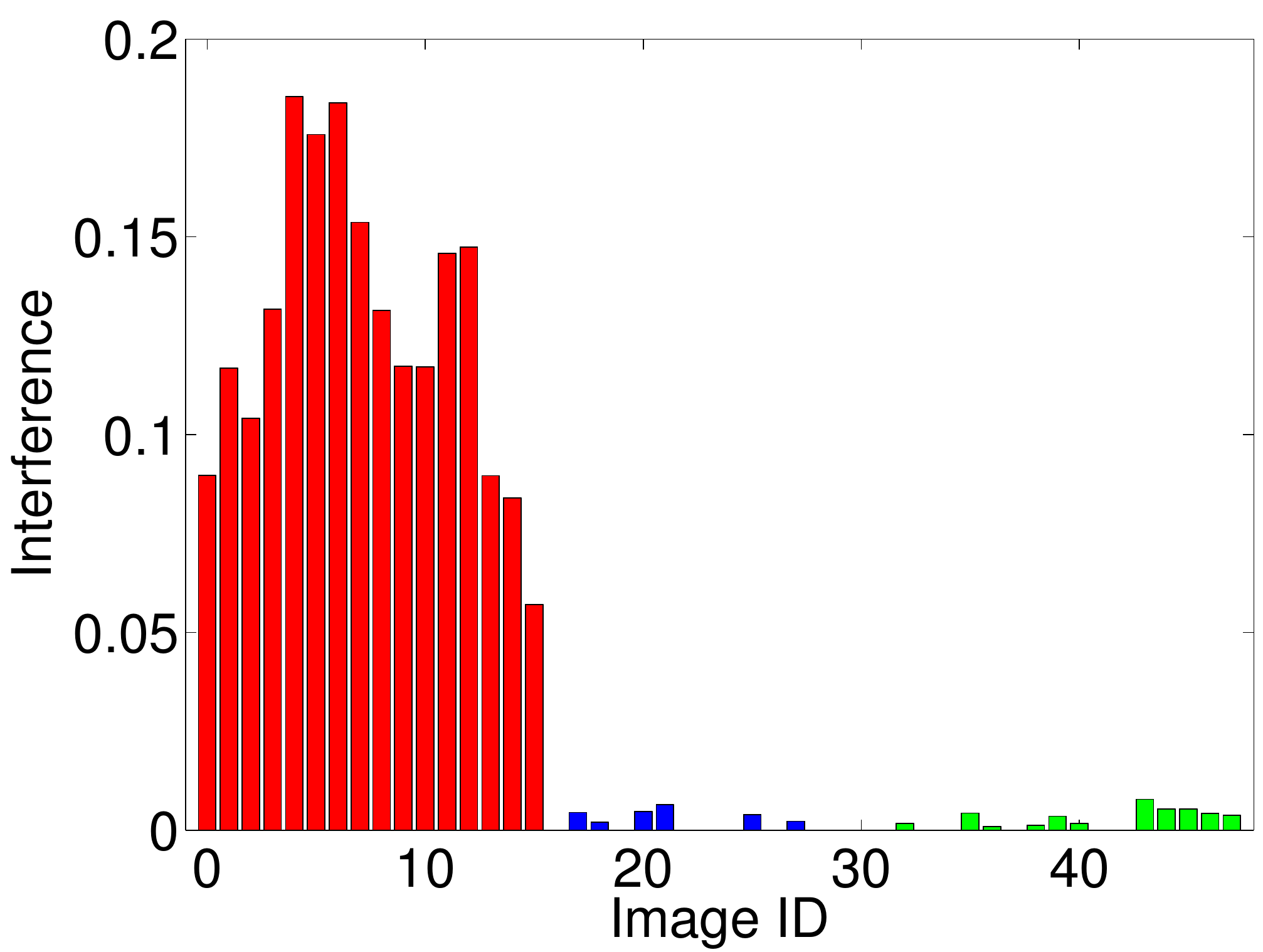} &
 \includegraphics[width=2in]{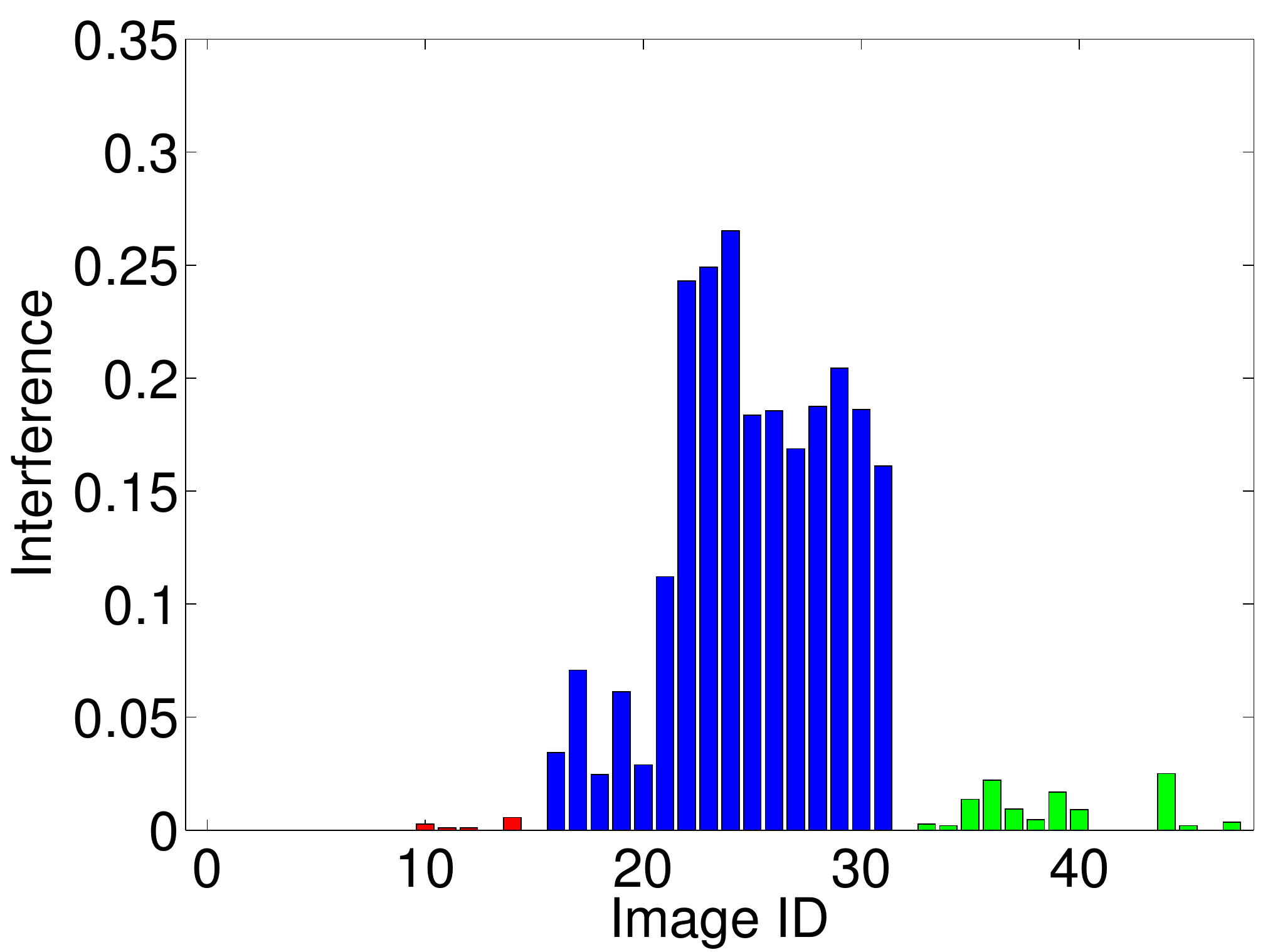} &
\\
 \includegraphics[width=2in]{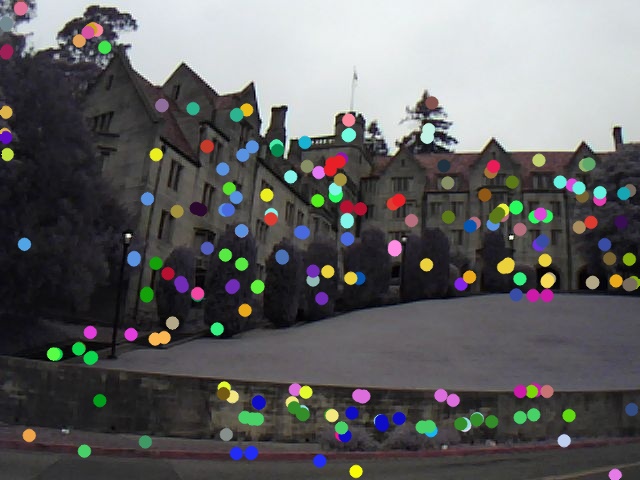} &
 \includegraphics[width=2in]{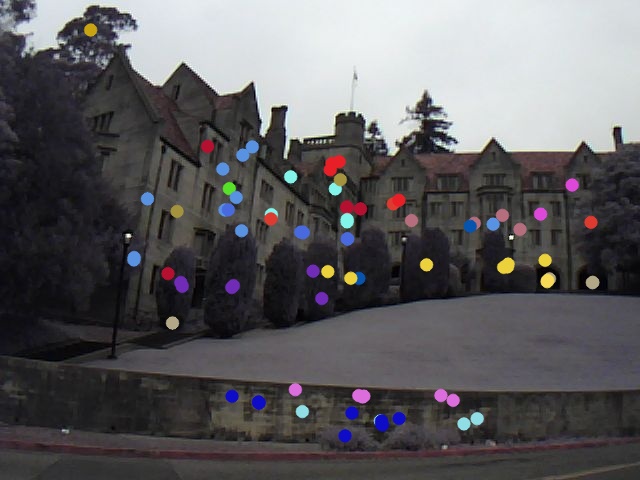} &
 \includegraphics[width=2in]{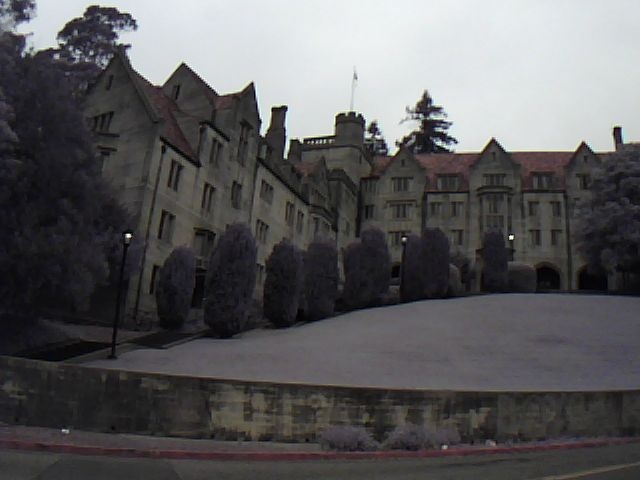} &
\\
 \includegraphics[width=2in]{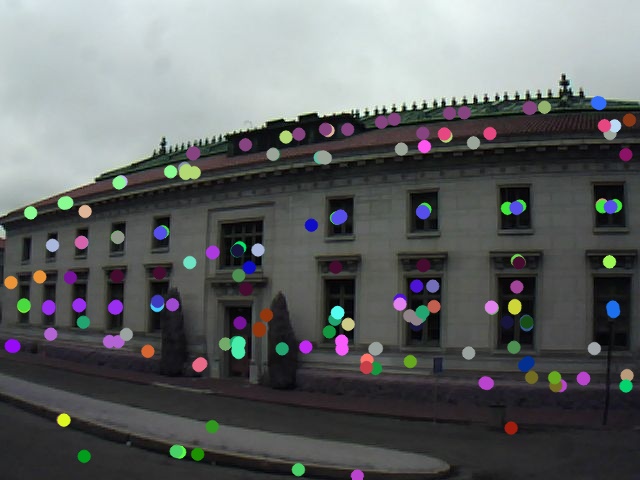} &
 \includegraphics[width=2in]{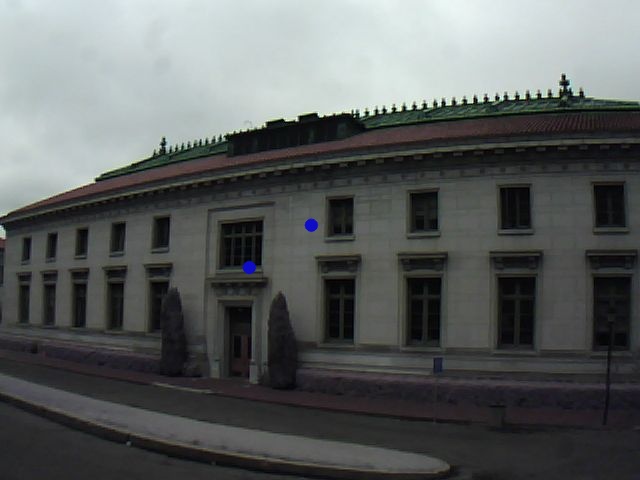} &
 \includegraphics[width=2in]{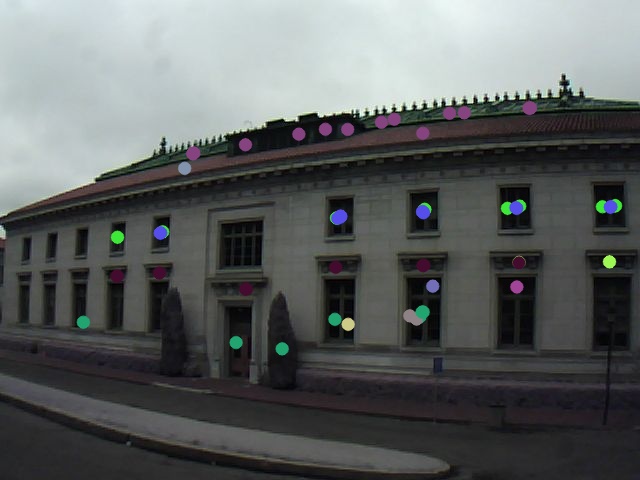} &
%
 \end{tabular}
 \caption{Different principal components select visual words prevalent in different categories.}
 \label{fig:correlationsOnBMW}
\end{figure}

For illustration purposes we limit our attention to just 2 topics; the message applies to more topics of course. The top row of Figure~\ref{fig:correlationsOnBMW} depicts the interference between sparse PCs $x^2, x^6$ and images belonging to three different topics/categories (red, blue and green bars). One can observe that indeed both sparse PCs have high interference with a single image topic. The next two rows show the same image three times; in the 1st column with all visual words, in the 2nd column with only those visual words selected by $x^2$ (i.e., in the set $\{j: x^2_j \neq 0\}$) and in the 3rd column only those visual words selected by $x^6$. Clearly, $x^2$ selects a substantial number of visual words in the second row image and does not select nearly any visual words in the third row image. The top image has high interference with $x^2$, while bottom image has low interference. The situation with $x^6$ is reversed. Indeed, the top image belongs to $S_2$, the topic attached to $x^2$, while the bottom image belongs to $S_6$.

 \begin{figure}[h!]
 \centering
 \subfigure[Projection of $\{h^i\}$ onto a random 3D subspace.]{\includegraphics[width=2.5in]{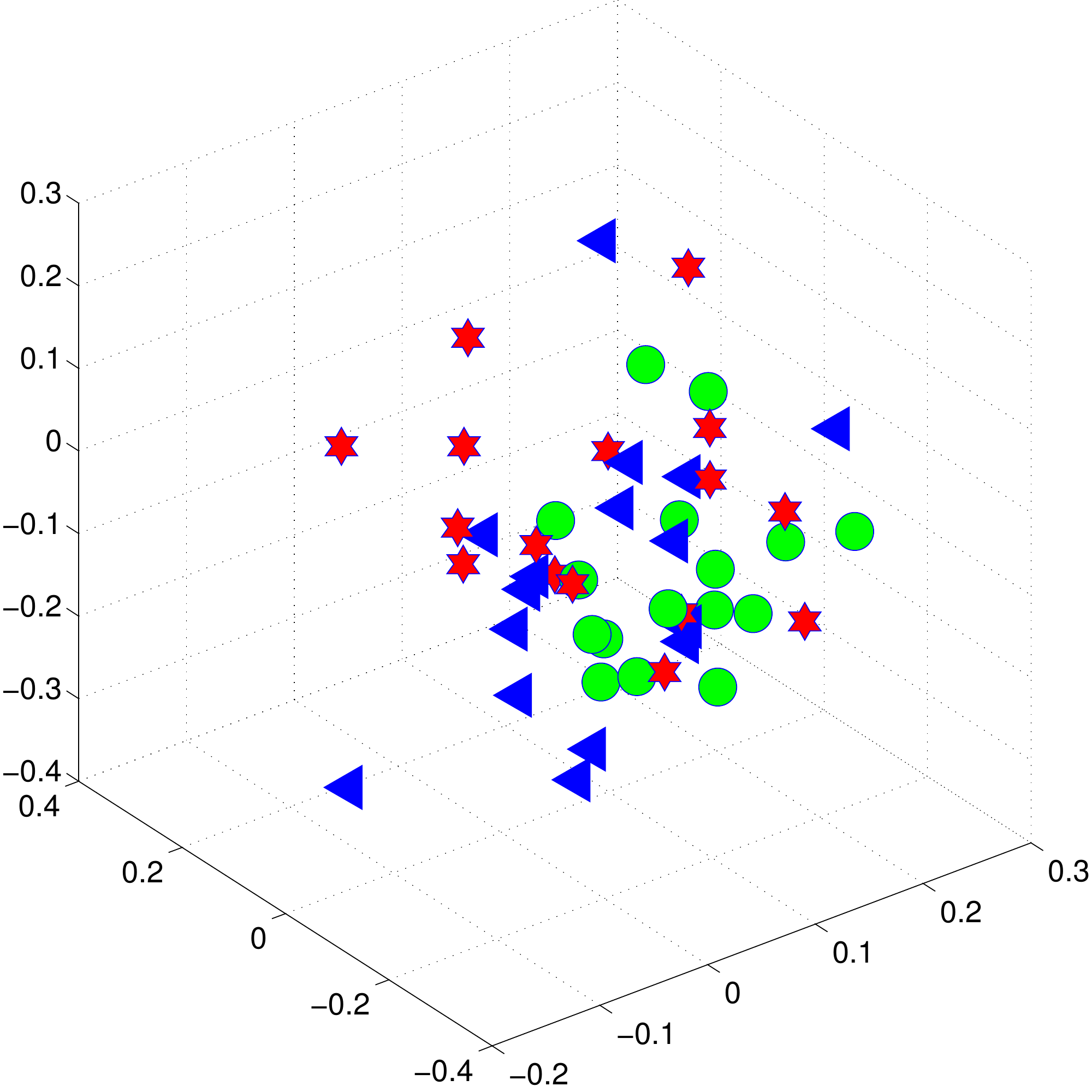}}
 \subfigure[Projection of $\{h^i\}$ onto the 3D space spanned by three sparse PCs.]{\includegraphics[width=2.5in]{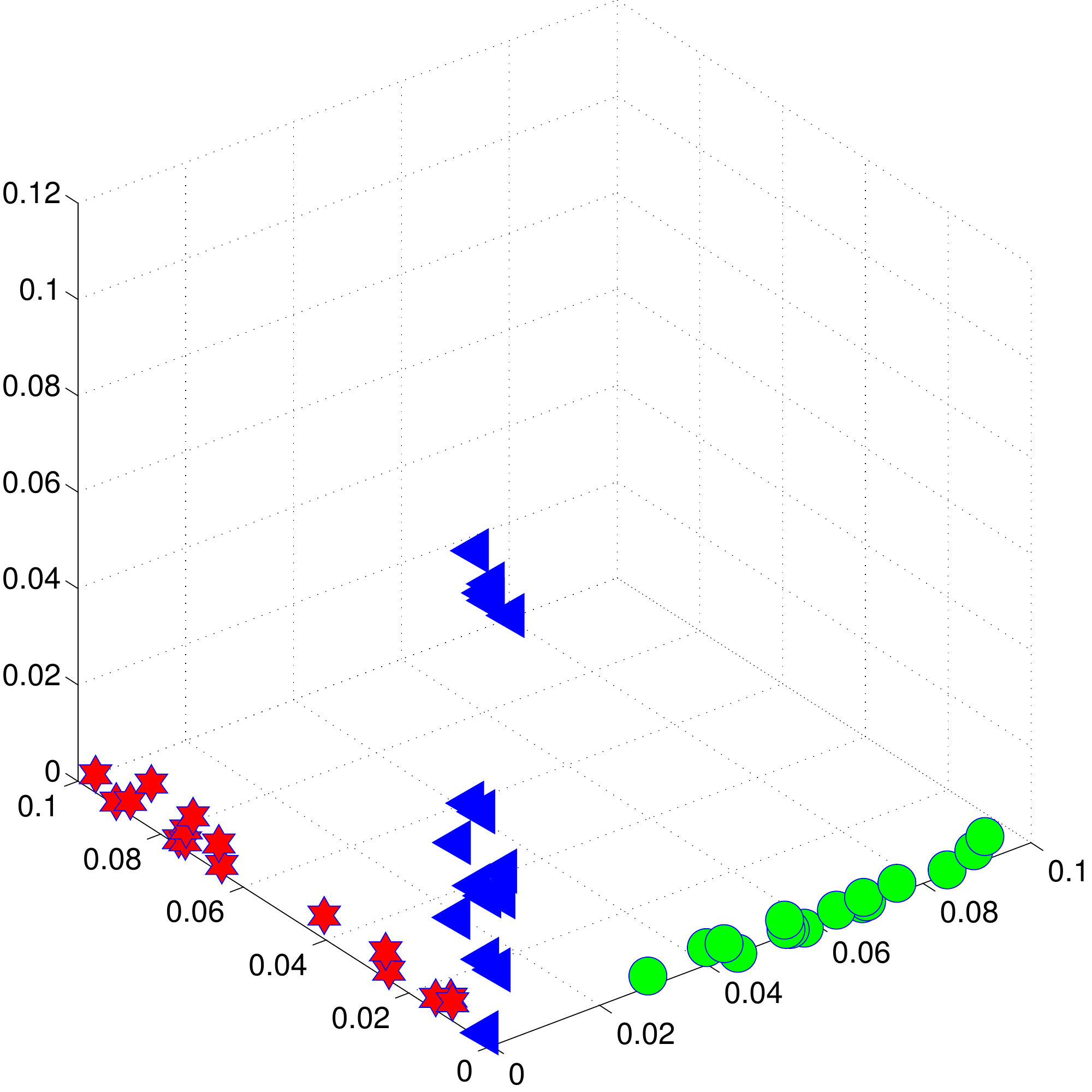}}
 \caption{Random vectors do not identify topics, sparse PCs do. }
 \label{fig:Projection}
  \end{figure}

\begin{figure*}
\centering

 \begin{tabular}{c}

\vtop{  
    \begin{tabular}{l}
        \includegraphics[width=2in]{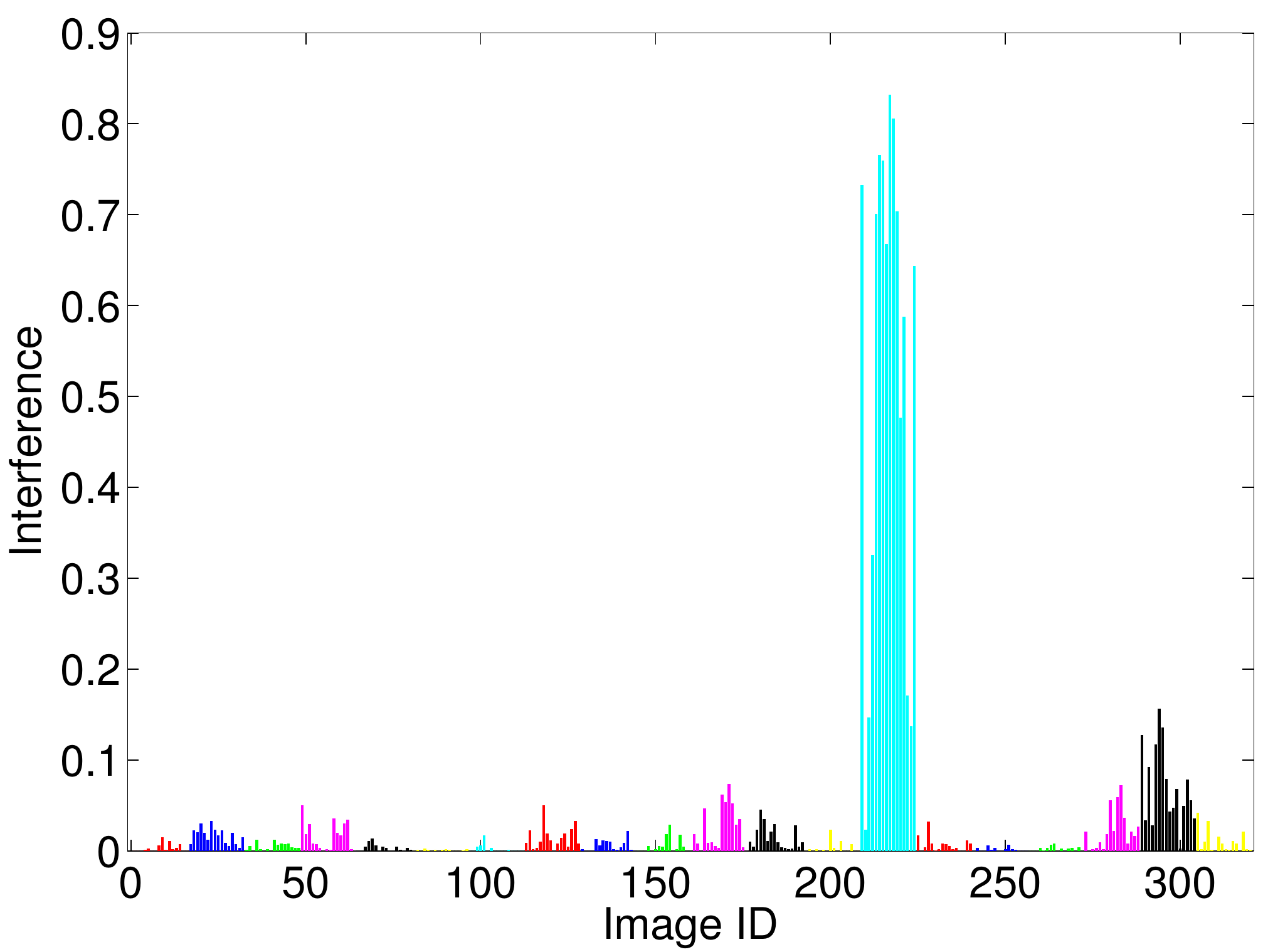}
    \end{tabular} 
    }%
\\
\vtop{
\begin{tabular}{l}
\includegraphics[width=1.2in]{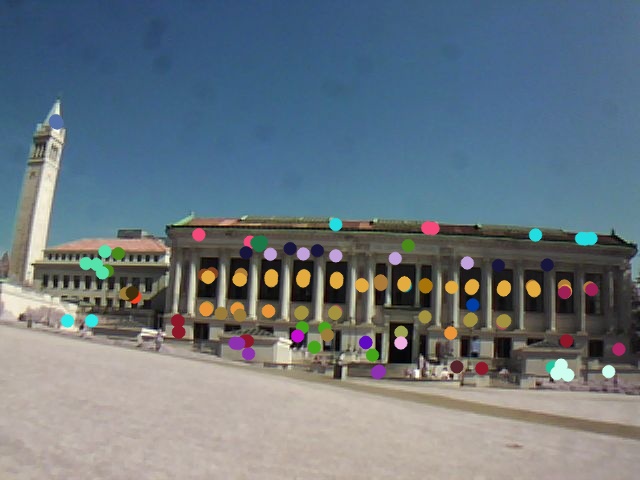}
\includegraphics[width=1.2in]{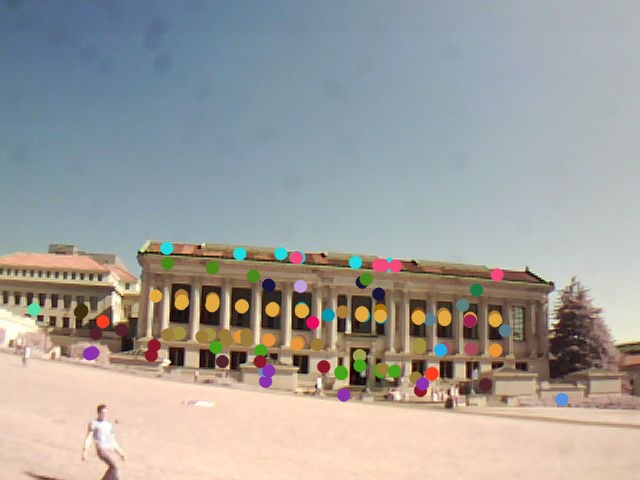}
\includegraphics[width=1.2in]{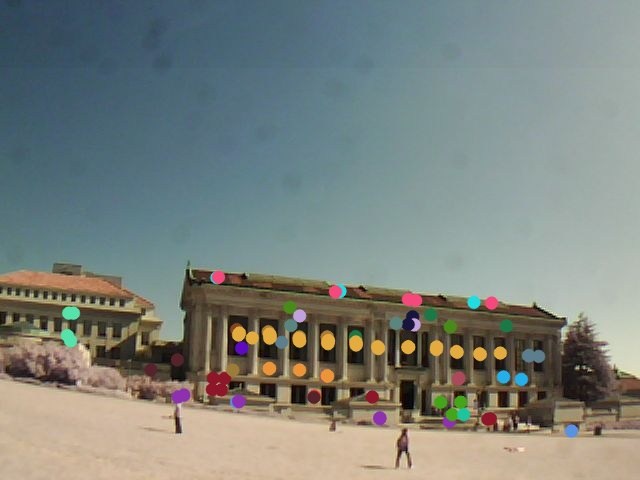}
\includegraphics[width=1.2in]{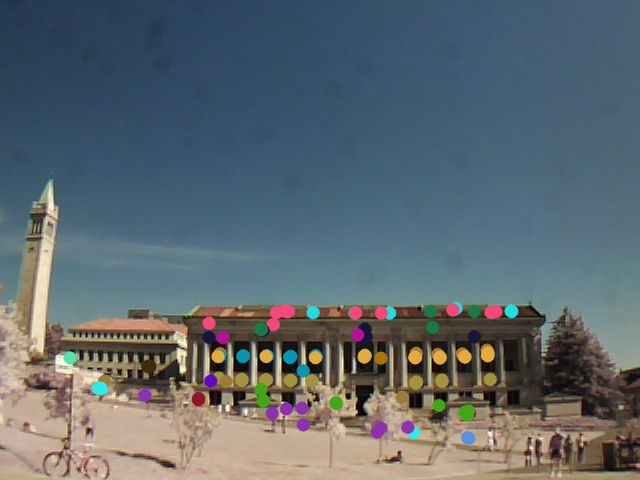}\\
\includegraphics[width=1.2in]{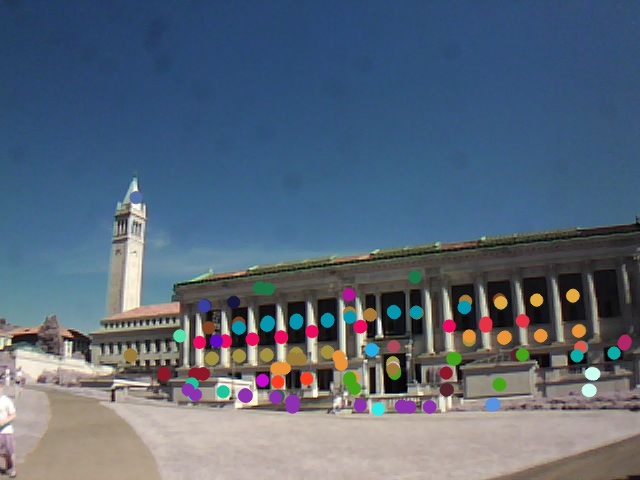}
\includegraphics[width=1.2in]{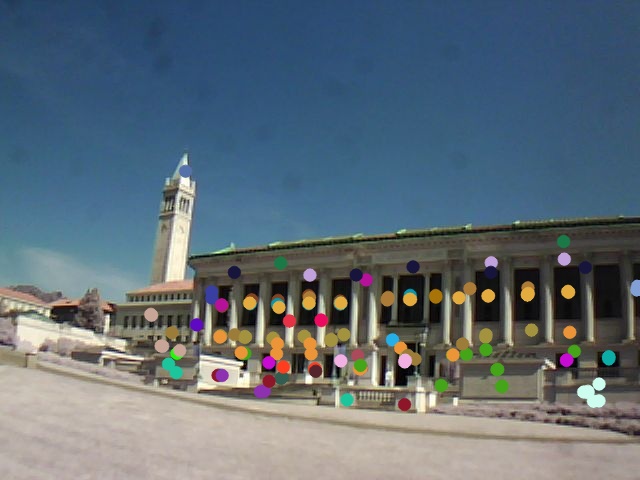}
\includegraphics[width=1.2in]{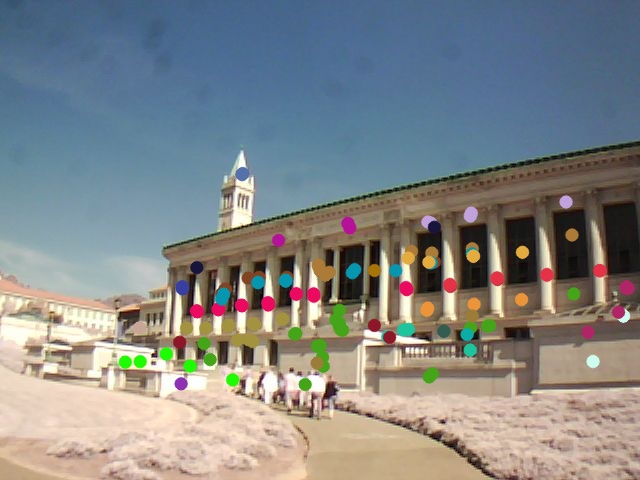}
\includegraphics[width=1.2in]{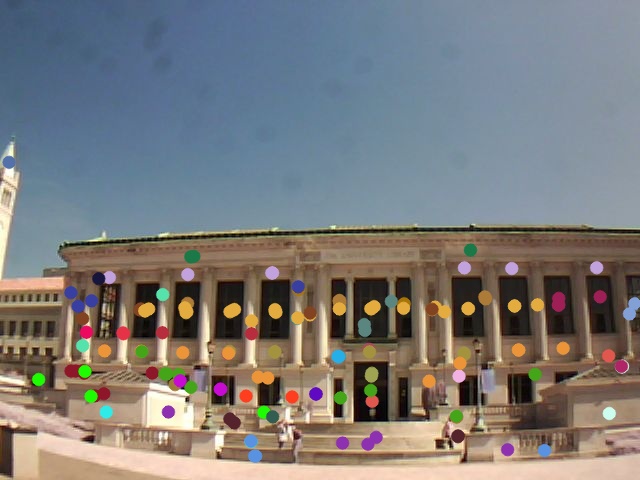}
\end{tabular}
}

\\

\vtop{  
    \begin{tabular}{l}
        \includegraphics[width=2.0in]{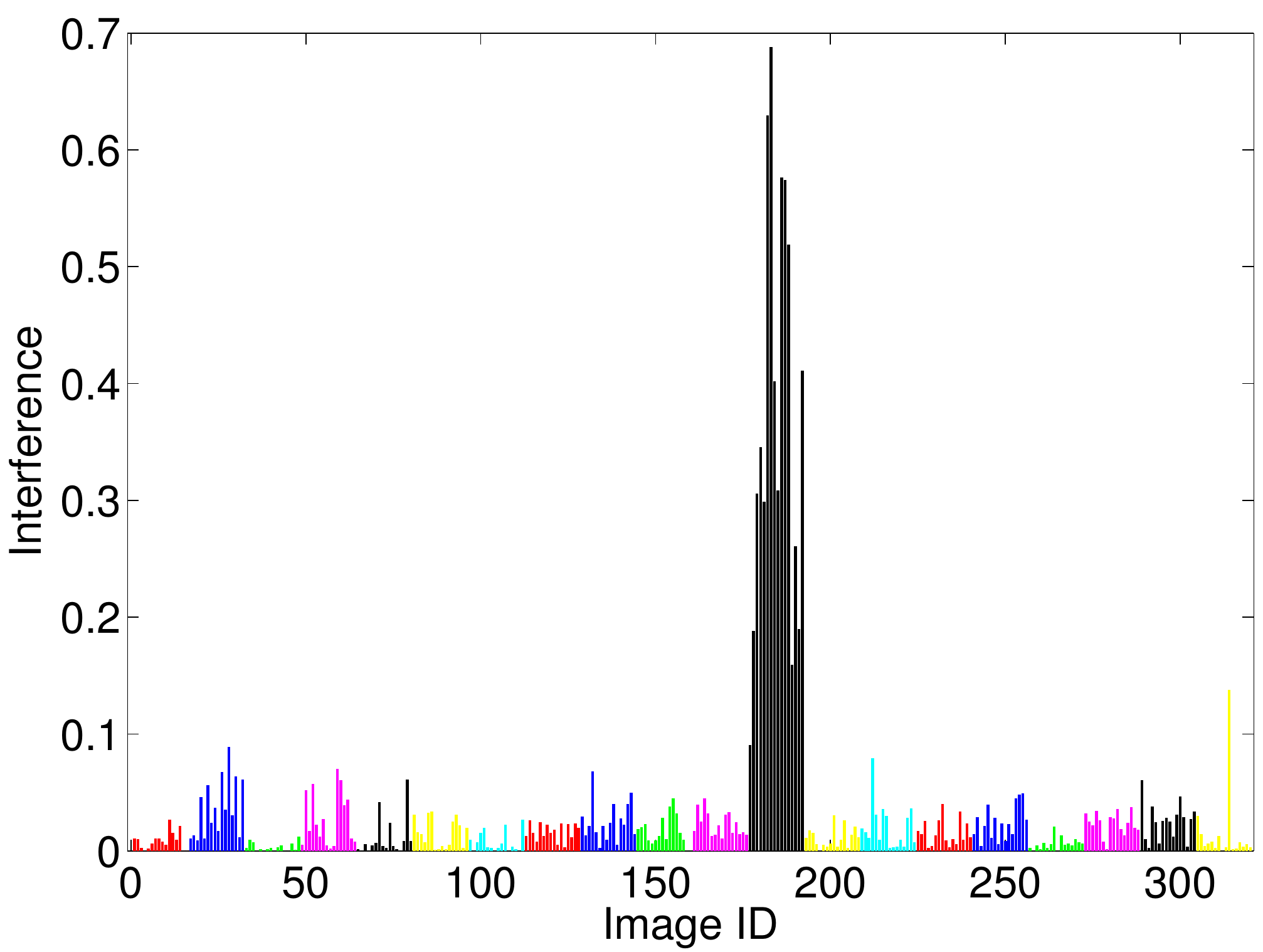}
    \end{tabular} 
    }%
\\
\vtop{
\begin{tabular}{l}
\includegraphics[width=1.2in]{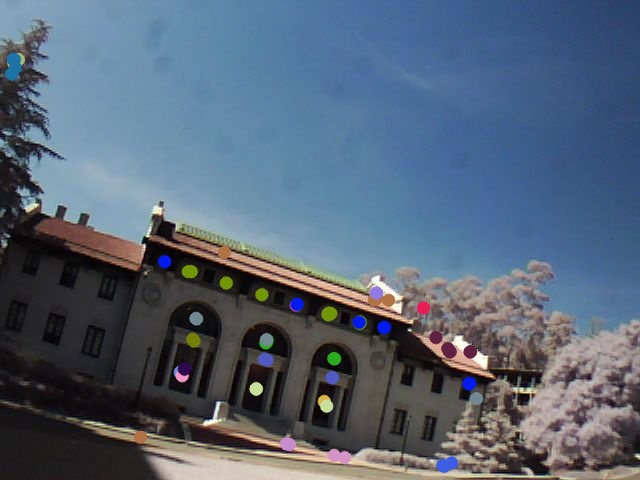}
\includegraphics[width=1.2in]{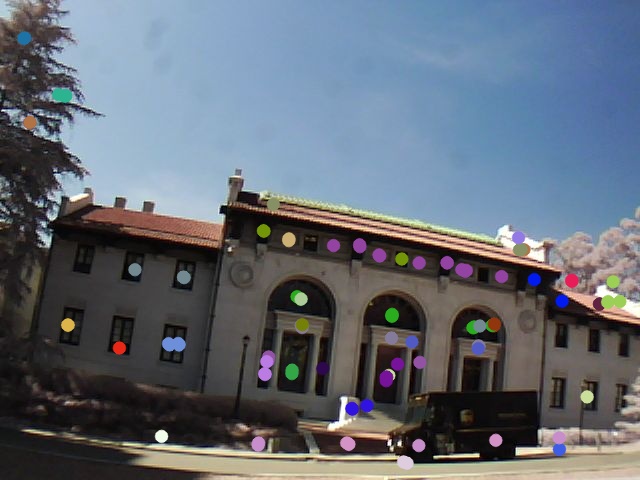}
\includegraphics[width=1.2in]{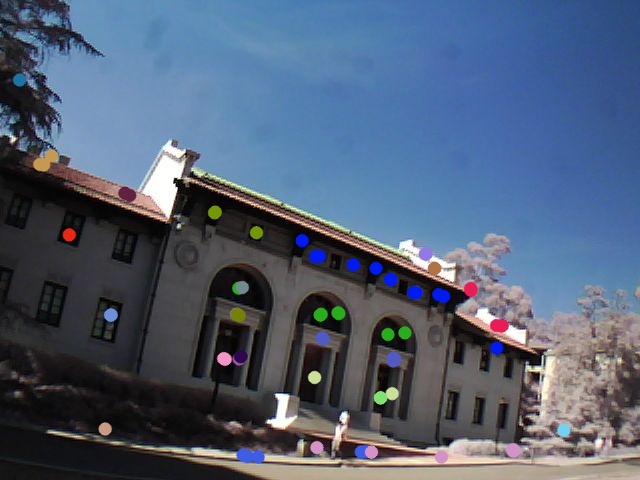}
\includegraphics[width=1.2in]{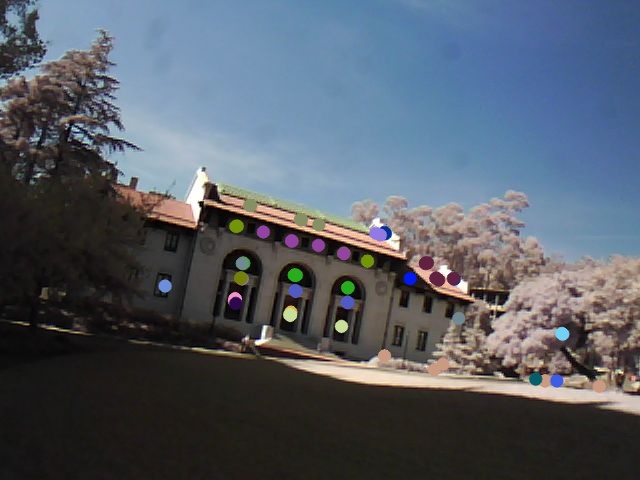}\\
\includegraphics[width=1.2in]{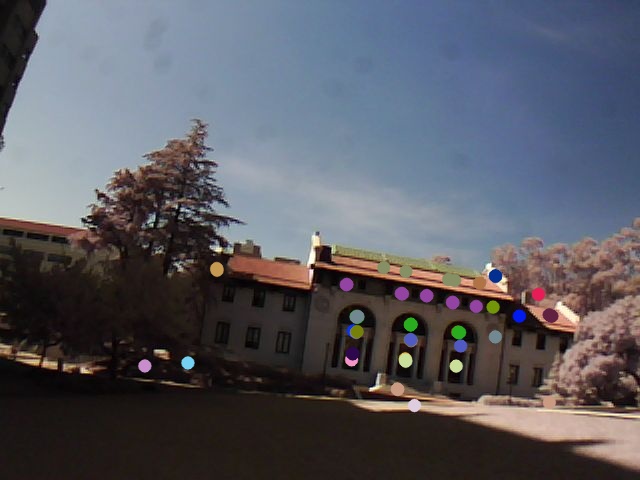}
\includegraphics[width=1.2in]{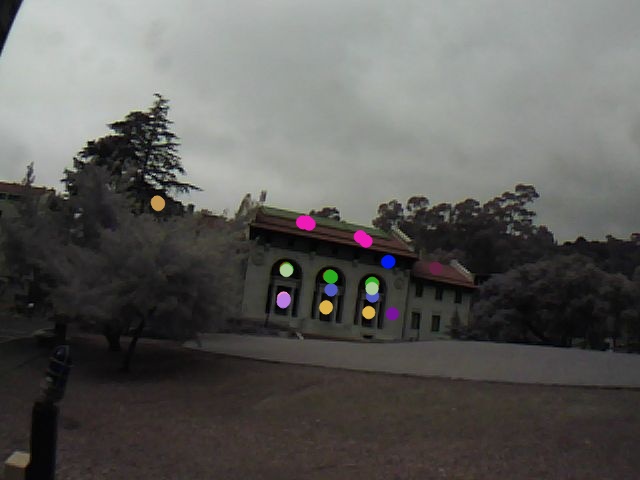}
\includegraphics[width=1.2in]{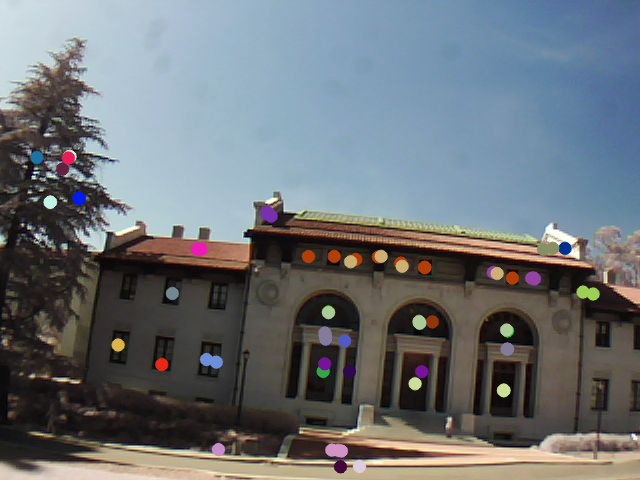}
\includegraphics[width=1.2in]{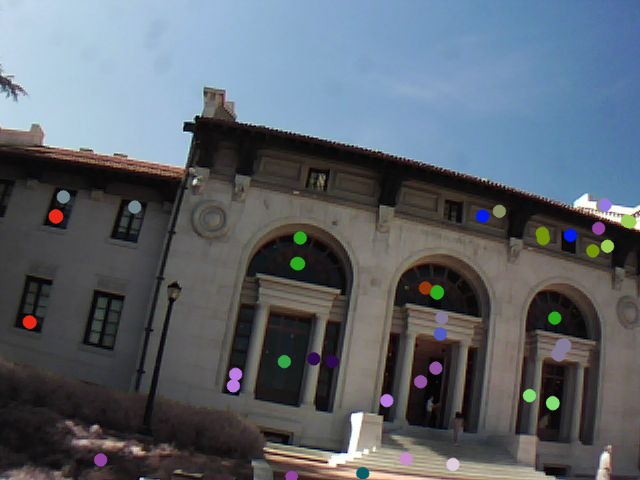}
\end{tabular}
}

  \end{tabular}
\caption{
The top part (plot + 8 images) correspond to sparse PC $x^1$, the bottom art (plot + 8 images) correspond to sparse PC $x^7$. The plots show the interferences of all images (horizontal axis) with the given PC. Images from the same category/topic (not known to our method!) are represented by the same color (but each color is used three times and each time it represents a different category/topic). For each PC we show the 8 images having the largest interference with it; these are the sets $S_1$ and $S_7$ for appropriate choice $\delta_1$ and $\delta_7$. One can observe that not only TOP-SPIN selects important features (visual words), but the method correctly identifies topics.}
\label{fig:bmw:interference}

\end{figure*}

In Figure~\ref{fig:Projection} we focus on the same three categories as in the previous test, but in this case we visualize them in 3D space, as in Figure~\ref{fig:motivation}. Because each image is represented by a $p=5,000$ dimensional vector ($h^i$), a na\"{\i}ve  approach for visualizing $h^i$ in 3D would be to project the vectors $h^i$ onto a random 3D subspace of $\R^p$ (Figure~\ref{fig:Projection}, left). No apparent separation of the images belonging to the three topics (represented by different color and marker) is present. However, if we project onto the space spanned by the PCs corresponding to the three topics, we can clearly see the images belonging to to different topics coalescing around different axes (Figure~\ref{fig:Projection}, right).

Let us now look at (a portion of) the actual output of TOP-SPIN for $k=7$, with a dictionary of size $p=5,000$ and $s=50$. Figure~\ref{fig:bmw:interference} depicts the sets $S_1$ and $S_7$ for $\delta_1$ and $\delta_7$ chosen so that $|S_1|=|S_7|= 8$. It is clear that the method is able to identify the categories.
We would like to stress that Sparse PCA is applied to the \emph{entire} training dataset, and that testing is done on different images. In contrast, the approach in \cite{Nikhil} presupposes the knowledge of the categories as Sparse PCA is applied to test images from each category. Moreover, as we shall see later, TOP-SPIN is able to also give better categorization accuracy.

\subsection{Category Prediction}\label{sec:bmwsection}

In this section we consider the problem of \emph{category prediction} (object recognition). While this is a different problem from the main focus of this paper: topic discovery, we will show that our framework can be also used to perform category prediction. Moreover, we demonstrate that our approach yields superior prediction accuracy results to the state of the art  \cite{Nikhil}.

Each image in the BMW dataset can be represented by a triple $(a,b,c)$, where $c$ is the category number (0--19), $b$ is camera number (0--4) and $a$ is the shot number (0--15). Let $M$ consist of all images with odd $a$ and $b = 2$ (i.e., $8$ images per category). The remaining images are partitioned into two groups: $T$ consisting of images with even $a$ and $b\neq 2$ ($32$ images per category) and $D$ (the rest; $40$ images per category). Finally, let $L$ be the set of all images with $b=2$. We will set aside $L$ for ``learning'', $M$ for ``matching'' and $T$ for ``testing''  as described below.

In the following we will describe and compare four methods, two from the literature (Baseline and NYS \cite{Nikhil}) and two new ones (Method 1 and Method 2), all of which perform the following category prediction task. Using images in $L$, learn a classifier which matches each image $i$ in the testing set $T$ to an image $m(i)$ in the matching set $M$. We then compute the prediction accuracy of each method defined as the percentage of images $i\in T$ for which  $i$ and $m(i)$ have the same category. The results are summarized in Table~\ref{tbl:bmw-comparison}; the description of the methods follows.

In all methods, a dictionary of $p=5,000$ visual words is first extracted from images in $L$, and then normalized histograms are computed for all images. In the Sparse PCA based methods we used 24AM for PC extraction.

\begin{enumerate}
\item \emph{Baseline.}  This classifier is given by $m(i) = \arg \min\{ \|h^{i} - h^{m}\|_1 : m \in M\}$. That is, we assign $i$ to image $m(i)$ whose histogram is closest to that of $i$ in $L_1$ norm.
\item \emph{NYS.} In \cite{Nikhil}, the authors  for each category $c$ form a matrix $A_c$ of normalized histograms corresponding to images in $L$ having category $c$, and then extract several sparse PCs of $A_c$. Let the  union of the supports of the PCs for every $c$ be $I_c$, and let $I=\cup_c I_c$.  The NYS classifier is given by $m(i) = \arg \min\{ \sum_{j \in I} |h^{i}_j - h^{m}_j| : m \in M\}$. This is similar to baseline, with the difference that only the important features ($I$) are used when computing the $L_1$ distance.
\item \emph{Method 1.} Here we propose a classifier similar to NYS with the exception that $I$ is obtained as the union of the supports of 160 $50$-sparse PCs of matrix $A_L$ whose rows are the normalized histograms of all images in $L$.
\item \emph{Method 2.} Here we compute 160 50-sparse PCs from $A_L$ and assign each PC to the image in $M$ with which it has the highest interference.  Then, when querying an image from $T$, we assign it to the PC with which it has the highest interference and through this, using the mapping just described, to an image in $M$.
\end{enumerate}

{\small

\begin{table}
\centering
 \caption{Category prediction accuracy of four methods; by category and total. Our approach improves on Baseline by 4\%.}
 \label{tbl:bmw-comparison}

\begin{tabular}{c|r|r|r|r}
 Cat. & Baseline & NYS & Method 1 & Method 2 \\ \hline \hline

0 & 100.00\%  & 100.00\%  & 100.00\%  &   100.00\% \\
1 & 90.62\%  & 93.75\%  & 90.62\% &    87.50\% \\
2 & 68.75\%  & 71.88\%  & 68.75\% &    87.50\%\\
3 & 96.88\%  & 96.88\%  & 100.00\% &   96.89\%\\
4 & 81.25\%  & 81.25\%  & 81.25\% &     100.00\%\\
5 & 100.00\%  & 100.00\%  & 100.00\% &  100.00\%\\
6 & 100.00\%  & 100.00\%  & 100.00\% &   81.25\%\\
7 & 81.25\%  & 81.25\%  & 84.38\% &    96.88\%\\
8 & 37.50\%  & 43.75\%  & 37.50\% &    81.25\%\\
9 & 40.62\%  & 46.88\%  & 46.88\% &    75.00\%\\
10 & 93.75\%  & 90.62\%  & 90.62\% &    81.25\%\\
11 & 100.00\%  & 100.00\%  & 100.00\% &   90.62\%\\
12 & 40.62\%  & 40.62\%  & 43.75\% &    37.50\%\\
13 & 100.00\%  & 100.00\%  & 100.00\% &  100.00\%\\
14 & 78.12\%  & 75.00\%  & 75.00\% &    78.12\%\\
15 & 96.88\%  & 93.75\%  & 96.88\% &  96.88\%\\
16 & 90.62\%  & 90.62\%  & 90.62\% &   78.12\%\\
17 & 100.00\%  & 100.00\%  & 100.00\% &  100.00\%\\
18 & 93.75\%  & 93.75\%  & 96.88\% &  100.00\%\\
19 & 100.00\%  & 100.00\%  & 100.00\% &   100.00\%\\

\hline \hline

Total & 84.53\% & 84.68\%   & 85.16 \% & {\bf 88.44\%}


\end{tabular}

\end{table}
} 

We can see from Table~\ref{tbl:bmw-comparison} that Method 2 is best (i.e., interference works better than $L_1$), then follows Method 1 (i.e., computing PCs using all of $A_L$ is better than computing PCs separately for each class), which is in turn superior to both NYS and Baseline. Method 2 outperforms Baseline by cca 4\%.

\emph{Remark 1:} Note that $T$ consists precisely of those images which do not have neither $a$ nor $b$ in common with any images in $M$. This is crucial as for every image in $D$ there is a (very) similar image in $M$ (one with the same $a$ but taken by a different camera $b$), which may skew the results. In fact, prediction accuracy on images from group $T$ is $69.2187\%$  and on images from group $D$ is $89.8750\%$, if we choose SURF descriptors and $p=1,000$, as in \cite{Nikhil}. This gap is present also when SIFT and $p=5,000$ is used, where the accuracy for group $T$ is $84.5313\%$ and for group $D$ is $96.6250\%$. This is the reason why we have discarded $D$ and used only $T$ for testing.

\emph{Remark 2:} Also note that a perfect comparison of our results with \cite{Nikhil} is not possible as all the data needed to reproduce the experiments exactly as in \cite{Nikhil} are not available to us. After implementing their method and setting all available options, we obtained Baseline prediction accuracy 80.69\%, whereas in \cite{Nikhil} the reported figure is 80.02\%. Moreover, we have decided not to use SURF descriptors but SIFT as this way we obtained better results.

\begin{figure}[ht!]
 \centering
 \includegraphics[width=2in]{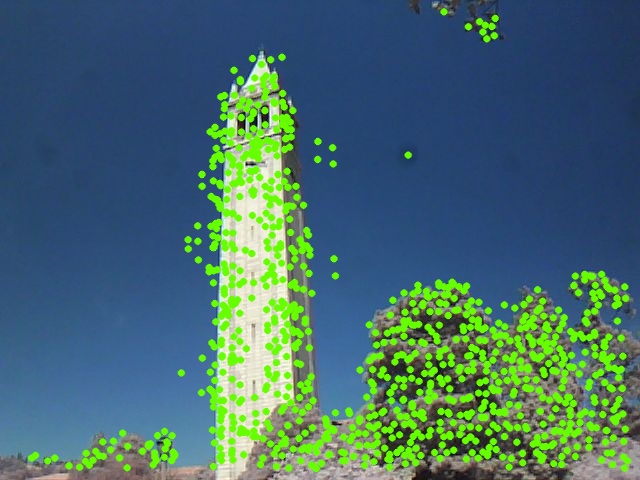}
 \includegraphics[width=2in]{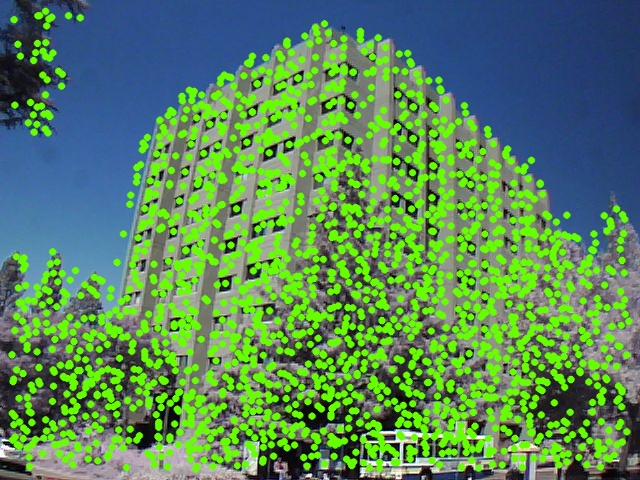}
 \includegraphics[width=2in]{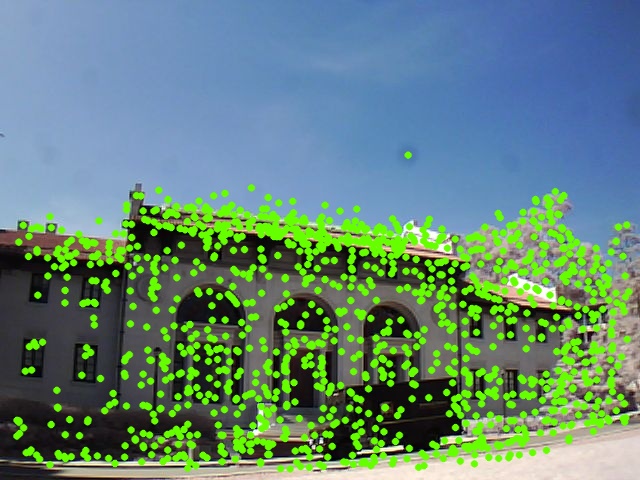}

 \includegraphics[width=2in]{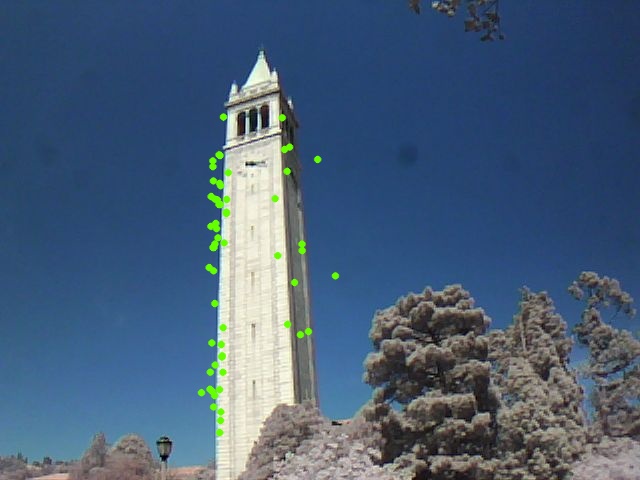}
 \includegraphics[width=2in]{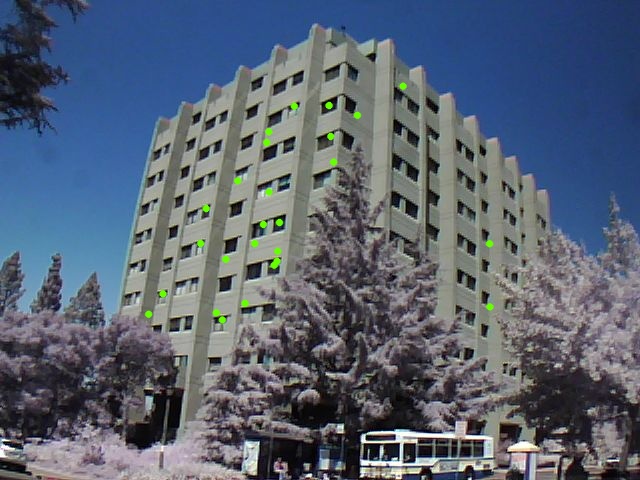}
 \includegraphics[width=2in]{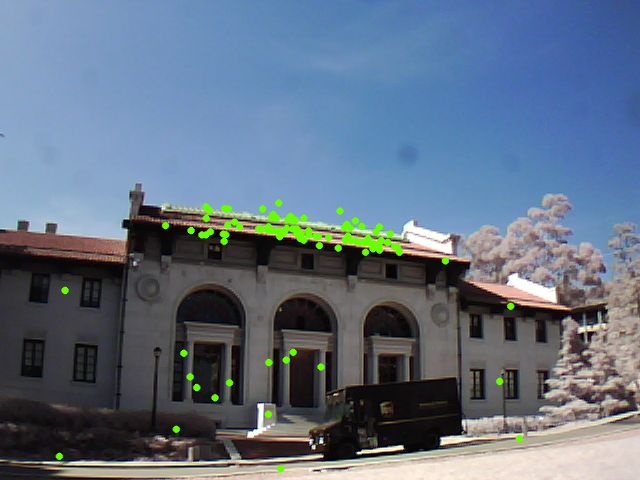}

 \includegraphics[width=2in]{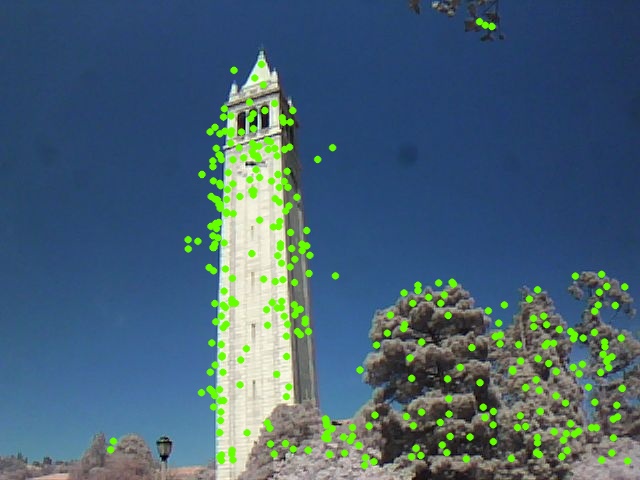}
 \includegraphics[width=2in]{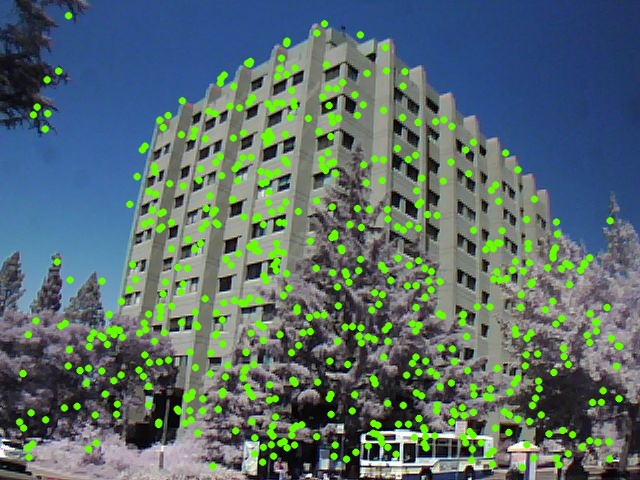}
 \includegraphics[width=2in]{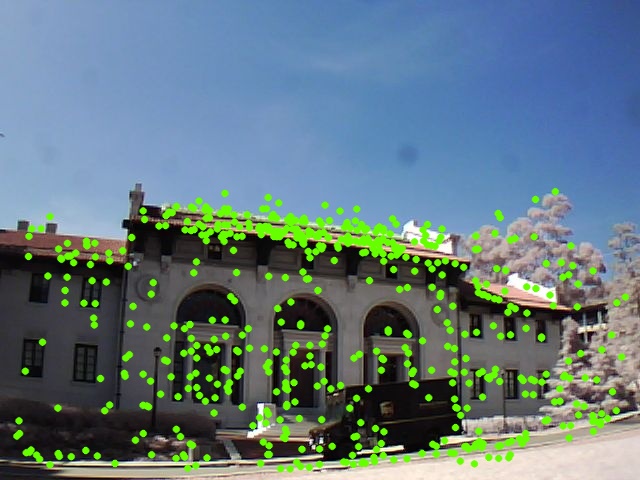}

 \includegraphics[width=2in]{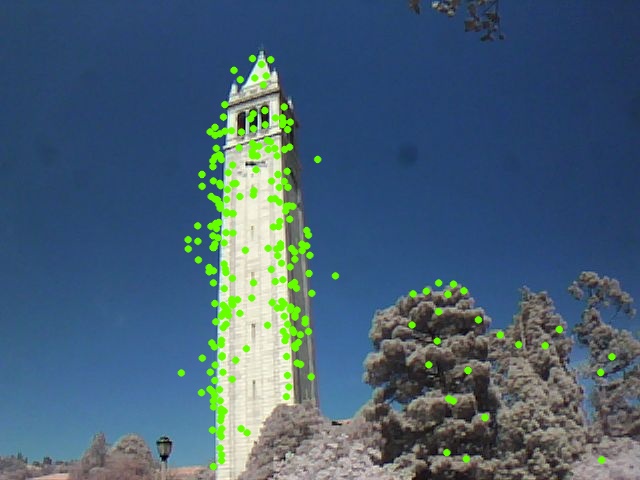}
 \includegraphics[width=2in]{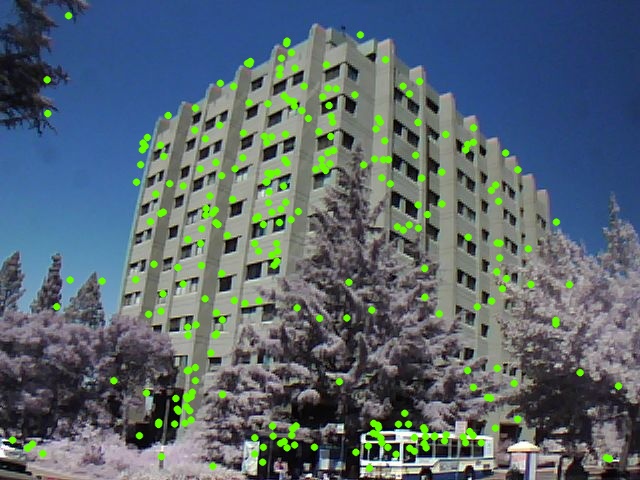}
 \includegraphics[width=2in]{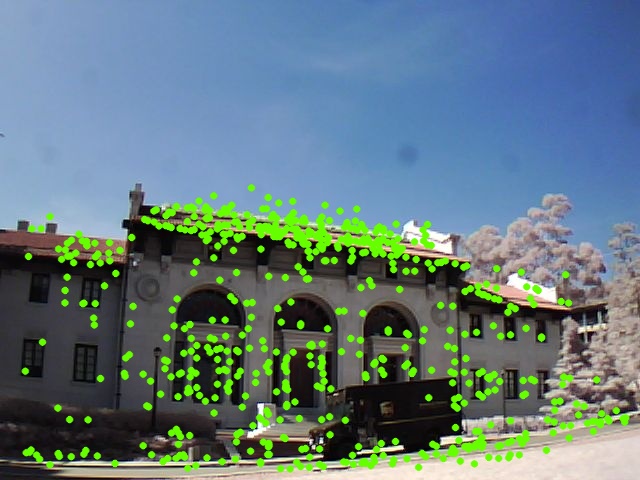}

 \caption{Features (visual words) selected by Baseline (first row), NYS per category = $I_c$ (second row), NYS in aggregate = $I$ (third row), and Method 1 / Method 2 (last row). }
 \label{fig:bmw-differentSupport}
\end{figure}

\begin{figure}
\centering
\includegraphics[width=5in]{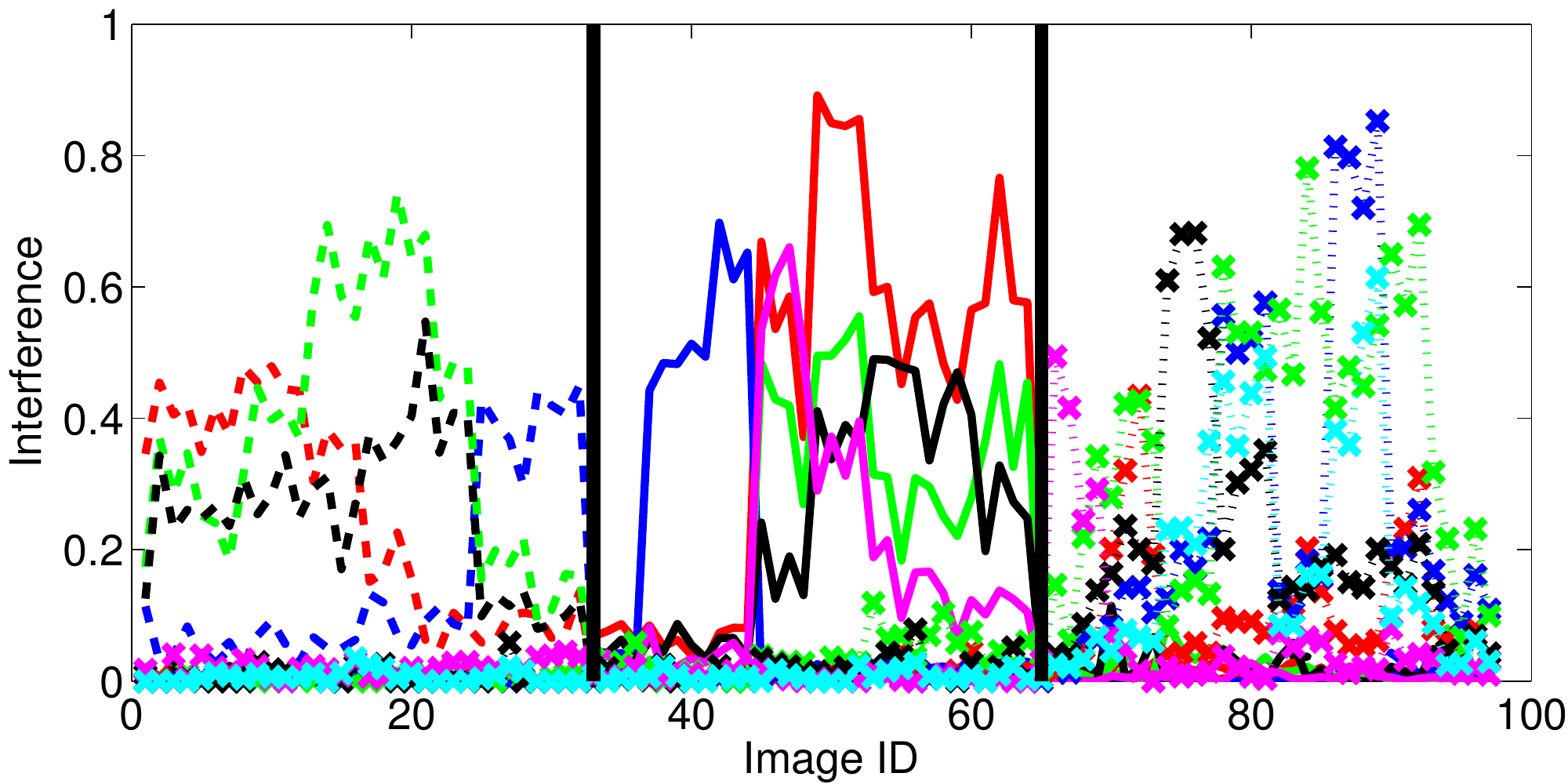}
\caption{160 PCs represented as 160 lines with unique formatting, and their interference with 96 images (32 test images from 3 categories). We see that each PC has high interference with a subset of images of a single category only, effectively selecting it. }
\label{fig:LAST}
\end{figure}

Let us now look at the features (visual words) selected by the four methods described above. In Figure~\ref{fig:bmw-differentSupport} we show 3 images from different categories. The first row shows all features in the dictionary appearing on these images. These are the features used by Baseline. The second row shows only features in $I_c$, for the three different values of category $c$ the three images belong to. In the third row we show the aggregate features $I = \cup_c I_c$. Finally, the last row show the features selected our approach (Method 1/Method 2). Note that we are able to achieve a better selection of features than NYS  (third row) \emph{without} the knowledge of image categories. The number of selected features for both NYS and PC1/PC2 was chosen to be the same for fairness of comparison purposes.

In Figure~\ref{fig:LAST} we give an additional insight into why Method 2 and TOP-SPIN work. The horizontal axis represents all test images belonging to three categories, CAT1, CAT2 and CAT3, the categories the  images in Figure~\ref{fig:bmw-differentSupport} belong to. That is, we consider $32 \times 3$ images. The first 32 images  correspond to CAT1; images 33--64 to CAT2 and images 65--96 belong to CAT3. Now, for \emph{all} 160 sparse PCs we plot a unique line in this plot, representing the interference of that PC with the images. For instance, the PC represented by the solid red line has high interference with images 45--64. Notice that all these images belong to CAT2. The PC corresponding to the solid blue line has high interference with 34--45, again a subset of images of CAT2. Note that neither the solid blue nor the solid red line have peaks in any of the other two regions/categories. This means that the PCs representing them effectively represent some object common to a subset of images in CAT2. The same is true for all other lines and the PCs they represent.

\section{Contributions}\label{sec:contributions}

We now summarize some of our main contributions:

\begin{enumerate}
\item We have developed an algorithm (TOP-SPIN) for solving the problem: topic discovery in a collection of unlabeled images.  Our algorithm applies Sparse PCA to identify co-occurred visual words that can be used as topic signatures.
\item  We have demonstrated on real datasets that TOP-SPIN is  able to discover topics and correctly assign images to the topics.
\item When used for category prediction, our framework gives higher accuracy than that of \cite{Nikhil}. Moreover, this is achieved without knowing what the categories are as sparse PCA is applied to data coming from all (test) images of all categories, not to (test) images of each category individually as in \cite{Nikhil}.
\item Our Sparse PCA solver is 3 or more order of magnitude faster than ALM.  It solves the Sparse PCA problem directly (i.e., not a relaxation), and unlike ALM, has direct control over the sparsity of the the PCs (via $s$).  Our Sparse PCA solver is parallel in nature and scalable to high-dimensions.
\end{enumerate}


\bibliographystyle{plain}
\bibliography{literature}


\end{document}